%% file: main.tex
\documentclass{article}
\usepackage{iclr2027_conference,times}
\usepackage{algorithm}
\usepackage[indLines=true]{algpseudocodex}
\usepackage{xspace}
\usepackage{booktabs}
\usepackage{multirow}
\usepackage{graphicx}
\usepackage{subcaption}

\usepackage{adjustbox}
\usepackage{makecell}
\usepackage{array}
\usepackage[table]{xcolor}
\usepackage{float}

\input{math_commands.tex}

\usepackage{hyperref}
\usepackage{url}
\usepackage{tabularx}

\newcolumntype{C}[1]{>{\centering\arraybackslash}m{#1}}
\renewcommand{\tabularxcolumn}[1]{m{#1}}

\definecolor{mainnodecolor}{HTML}{0072B2}
\definecolor{discoverycolor}{HTML}{D55E00}
\definecolor{refinementcolor}{HTML}{009E73}

\newcommand{\MainNode}[1]{%
  \textcolor{mainnodecolor}{\textbf{M:#1}}%
}
\newcommand{\DiscoveryNode}[1]{%
  \textcolor{discoverycolor}{\textbf{D:#1}}%
}
\newcommand{\RefinementNode}[1]{%
  \textcolor{refinementcolor}{\textbf{R:#1}}%
}
\newcommand{\AGBUNode}[1]{%
  #1\textsuperscript{\(\dagger\)}%
}
\newcommand{\PathArrow}{%
  \hspace{2pt}\(\rightarrow\)\allowbreak\hspace{2pt}%
}

\newcolumntype{Y}{>{\centering\arraybackslash}X}

\definecolor{bestrow}{RGB}{252,232,232}

\newcolumntype{V}{!{\color{black}\vrule width 0.35pt}}

\newcommand{\workname}{SkillVine\xspace}

\iclrfinalcopy 

\title{\workname: Agent Skill Evolution via Branching Exploration}

\author{
\textbf{
Kaiwei Liu$^{1}$%
\thanks{Equal contribution.\quad
\textsuperscript{\(\ddagger\)}Corresponding author.}%
\hspace{0.4em}%
\thanks{This work was done while Kaiwei Liu was an intern at Huawei Noah's Ark Lab.},
Jiqian Dong$^{2}$\footnotemark[1],
Liran Dong$^{1}$,
Shuai Mao$^{2}$,
Mingming Zhao$^{2}$} \\
\textbf{
Bufang Yang$^{1}$,
Jie Chuai$^{2}$,
Zhitang Chen$^{2}$,
Guoliang Xing$^{1}$,
Zhenyu Yan$^{1}$\footnotemark[3]} \\[4pt]
$^{1}$The Chinese University of Hong Kong
\quad
$^{2}$Noah's Ark Lab, Huawei Technologies \\[4pt]
\texttt{\{kaiweiliu,bfyang\}@link.cuhk.edu.hk}\quad
\texttt{dl123@ie.cuhk.edu.hk} \\
\texttt{\{glxing,zyyan\}@cuhk.edu.hk}\quad\texttt{\{chuaijie,chenzhitang2\}@huawei.com}\\
\texttt{\{dong.jiqian,mao.shuai677,zhaomingming9\}@huawei.com}
}

\begin{document}

\maketitle
\fancyhead{} 
\renewcommand{\headrulewidth}{0pt} 

\input{sec/abstract}
\input{sec/introduction}

\input{sec/related_work}

\input{sec/method}

\input{sec/experiments}
\input{sec/conclusion}
\input{sec/ai_statement}
\input{sec/ref}

\clearpage
\input{sec/appendix}

\end{document}

%% file: math_commands.tex
\usepackage{amsmath,amsfonts,bm}

\def\eqref#1{equation~\ref{#1}}

\def\1{\bm{1}}

\DeclareMathAlphabet{\mathsfit}{\encodingdefault}{\sfdefault}{m}{sl}
\SetMathAlphabet{\mathsfit}{bold}{\encodingdefault}{\sfdefault}{bx}{n}



%% file: sec/abstract.tex
\begin{abstract}
\vspace{-1em}
Agent skills encapsulate reusable procedural knowledge that enables LLM agents to perform tasks, and they can be improved automatically using trajectories from interactions with the environment. This is the classic problem of skill evolution. Existing approaches predominately follow a linear evolution paradigm, in which updates are sequentially applied to the latest skill-library version. As a result, they inevitably fall into local optima, leaving many promising evolution paths unexplored. We propose \workname, an automatic skill-evolution framework that formulates skill evolution as a graph search problem and employs a branching exploration strategy. Equipped with a trunk-branch collaborative searching mechanism, an intelligent parent-node selector, and an adaptive-granularity update rule, \workname achieves a balance between exploration and exploitation. We evaluate \workname on 5 benchmarks with two LLMs. Results show that SkillVine discovers better skill-library versions along branches than along the linear trunk and achieves the best test performance in nine of ten benchmark–model combinations.

\end{abstract}

%% file: sec/introduction.tex
\section{Introduction}

Large Language Model (LLM) agents are increasingly used in software engineering~\citep{yang2024swe, jimenez2024swe, qian2024chatdev}, automated scientific research~\citep{team2025tongyi, openai2025deepresearch, lu2024ai, chan2025mle}, web interaction~\citep{googledeepmind2025computeruse, openai2025cua, zhou2024webarena, yao2022webshop}, and embodied intelligence~\citep{bolton2025sima, wang2023voyager}. These applications require both general reasoning and specialized knowledge, yet putting extensive procedural knowledge directly in prompts is constrained by fixed context windows. Agent Skills~\citep{anthropic2025skills} address this limitation by packaging reusable procedures as modular, dynamically loadable executable units. However, static skill libraries cannot readily adapt to changing environments or tasks beyond their initial coverage, motivating research on self-evolving agents~\citep{self-evolving-agent-survey, dynamic_agent_survey} and automatic skill acquisition from interaction trajectories~\citep{trace2skill, auto-skill, anything2skill}. Yet simply adding more skills does not always improve performance: expanding a library from a small collection to 202 skills can reduce performance by up to 21\%, with skill shadowing, in which agents increasingly select inappropriate skills as the library grows, accounting for up to 68\% of this degradation~\citep{more-skills-degrade-performance}. Lifelong skill evolution must therefore encompass not only acquisition but also systematic refinement, pruning, and adaptation to sustain agent competence and safety in non-stationary environments.

Recent work has moved beyond one-time skill generation toward lifelong skill evolution. Existing approaches use reinforcement learning or model self-reflection to optimize either the agent's parameters~\citep{wu2025evolver, xia2026skillrl} or its external skill library~\citep{ouyang2026skillos, yang2026skillopt, skilloptlite_github}. Most approaches, however, follow a \textbf{linear evolution paradigm} in which every update is applied sequentially to the latest skill library state. This paradigm introduces three limitations. 
First, as the skill library evolves along a single forward path, useful content from earlier versions may be overwritten, especially when new tasks differ greatly from previous ones.
Second, each update builds sequentially on all previous updates, and some updates may lead evolution into a local optimum. Without version history, the system cannot undo these updates and explore alternative paths. For example, an update may add a rule that helps with current tasks but limits performance on later tasks. Under a linear evolution paradigm, the system cannot return to the version before this rule was added and explore a different path. 
Third, the linear paradigm forces the skill library to evolve along a single path, preventing parallel exploration and systematic comparison of different skill variants’ performance. 
To overcome these limitations, we propose \textbf{\workname}, a framework that enables agent skill evolution through \textbf{branching exploration}.

To enable branching exploration, \workname organizes skill-library versions into a \textbf{version graph} that preserves the complete content of each version. Each node stores a skill-library version and its validation results, while directed edges record parent-child relationships between nodes, allowing the system to return to any earlier versions and explore new paths from them. 
\workname introduces three key mechanisms. 
First, it uses a \textbf{trunk-branch update strategy}. The trunk attempts an incremental update on every training batch, allowing useful updates to accumulate continuously along it. Branches grow from historical versions to explore alternative paths that provide capabilities the trunk lacks.
Second, \workname uses a \textbf{complementarity-aware node selector} to choose historical skill-library versions as starting points for branch evolution, favoring those whose capabilities complement the trunk. It scores each version based on its validation accuracy, how many candidate child versions it has produced, and how its capabilities complement the trunk.
Third, it uses \textbf{adaptive granularity} for branch updates to maintain an appropriate optimization step size based on whether the trunk accepts its update. This reduces the risk that noisy or misleading information in a training batch leads to harmful skill updates. 
Overall, the design of \workname automatically balances performance, exploration, and capability complementarity within a limited optimization budget.

We evaluate \workname with two models on five benchmarks covering spreadsheet manipulation, mathematical reasoning, and question answering.
\workname outperforms six baselines in nine of ten model–benchmark combinations. With DeepSeek-V4-flash, it leads on all five benchmarks. 
Moreover, every best final skill library is discovered through a branch rather than the trunk, demonstrating the value of branching beyond linear evolution.

Overall, our main contributions can be summarized as follows:
\begin{itemize}
    \item We propose \workname, a framework for agent skill evolution through branching exploration. It preserves historical skill-library versions and combines continuous improvement along a trunk with exploration of alternative paths through branches.
    \item We design a complementarity-aware node selector and an adaptive branch update strategy. The selector selects historical versions by jointly considering their performance, how often they have been expanded, and how their capabilities complement the trunk, while the update strategy adjusts branch update granularity based on the trunk's update outcome to help filter out harmful updates.
    \item We conduct extensive experiments on five benchmarks with two base LLMs. 
    \workname achieves the best test performance in nine of ten benchmark--model combinations. 
    Analysis shows that branches discover better skill library versions than those found along the trunk.
\end{itemize}

%% file: sec/related_work.tex
\section{Related Work}
\vspace{-0.5em}

\subsection{Lifelong Evolution of Skills}
\vspace{-0.5em}

The lifelong management of agent skills has recently gained traction as a pivotal research frontier within self-evolving agentic systems. Survey \citet{dynamic_agent_survey} formalizes this domain through an eight-stage lifecycle, spanning evidence collection, proposal, validation, organization, retrieval, maintenance, distillation, and governance; and a ten-operator taxonomy (e.g., Add, Refine, Merge, Prune) designed to systematically govern skill trajectories. A growing body of contemporary literature operationalizes this conceptual architecture across specific domains, such as automatic skill curation \citep{trace2skill, auto-skill, anything2skill, muse-autoskill, ouyang2026skillos, xia2026skillrl}, test-time skill optimization \citep{yang2026skillopt, skilloptlite_github}, and continuous lifelong management \citep{life-skill, ell, skill-lens, zhang2026arenaflow}. Collectively, these contributions signal a paradigm shift in the field: moving beyond naive skill storage toward scalable, verifiable, and safe lifelong evolution.

\subsection{Branching Search and Exploration}
\vspace{-0.5em}

Branching search has long been used to balance exploration and exploitation in large decision spaces.
Monte Carlo Tree Search (MCTS) and its UCT variant~\citep{kocsis2006bandit, browne2012survey} perform value-guided expansion over search trees.
Go-Explore~\citep{ecoffet2019go} explicitly archives promising states, returns to them, and then explores further.
Population-Based Training~\citep{jaderberg2017population} periodically copies high-performing training states and perturbs them to continue exploration.
Recent LLM agents have also incorporated branching search to improve reasoning, acting, and tool-use trajectories.
LATS~\citep{zhou2023language} integrates MCTS with LLM-powered agents, value functions, and self-reflection,
while subsequent work applies tree search to web agents~\citep{koh2024tree} and agentic AutoML~\citep{liang2026mcts}.
In parallel, MCTS has also been used to synthesize reasoning trajectories or process supervision signals for LLM training~\citep{luo2024improve, guan2025rstar}.
Unlike these methods, which branch over action, reasoning, or tool-use trajectories, our method branches over skill-library versions. Each node represents a complete skill-bank snapshot, and each edge corresponds to a parent-child relationship.

%% file: sec/method.tex
\section{\workname}
\vspace{-0.5em}

This section formulates skill library evolution as a budget-constrained optimization problem and introduces \workname, which performs branching exploration over skill-library versions using a recoverable version graph.

\subsection{Problem Formulation}
\vspace{-0.5em}

We formulate skill evolution as a sequential optimization problem that maximizes agent performance on a validation distribution subject to a fixed budget.
Let $S_v=\{s_{v,1},\ldots,s_{v,n_v}\}\in\mathcal{S}$ denote the skill library at version node $v$, where $n_v$ is the number of skills in that version. Each skill $s_{v,i}$ is a reusable procedural unit (e.g., a guideline or executable script) that an LLM agent $\pi$ can use to complete a task. 
At each evolution step $t$, the agent interacts with a stream of tasks sampled from a distribution $\mathcal{D}_t$, generating  trajectories $T_t=\{\tau_{t,1},\ldots,\tau_{t,m}\}\in\mathcal{T}$. 
Each trajectory $\tau_{t,j}=(x_{t,j},a_{t,j},y_{t,j})$
comprises the input prompts $x_{t,j}$, the executed action sequence $a_{t,j}$ (including skill invocations), and the outputs $y_{t,j}$. 
These trajectories are then aggregated and summarized to produce candidate skill updates (i.e., additions, refinements, or deletions). 
Formally, an update operator $U:\mathcal{S}\times\mathcal{T}\rightarrow\mathcal{S}$ maps a selected skill library version $S_v$ and the interaction trajectories $\mathcal{T}_t$ collected with that version at training step $t$ to a candidate child version: $S_{v'}=U(S_v,\mathcal{T}_t)$.

The objective of skill evolution is to identify the optimal skill library $S^*$ that maximizes expected accuracy on a validation distribution, subject to a constrained training budget. 
For each candidate update generated from a task stream, we make a binary decision: either discard the update or apply it to a specific skill library version, thereby systematically exploring alternative evolutionary paths.
Let $\mathcal V_B\subseteq\mathcal S$ denote the set of skill library versions explored within training budget \(B\). The corresponding optimization problem is formally defined as:
\begin{equation}
S^*=\arg\max_{S\in\mathcal{V}_B}
\operatorname{Acc}(\pi,S;\mathcal{D}_{\mathrm{val}}),  
\end{equation}
where \(\operatorname{Acc}(\pi,S;\mathcal D_{\mathrm{val}})\) denotes the expected accuracy of agent \(\pi\) equipped with skill library \(S\) on the validation distribution \(\mathcal D_{\mathrm{val}}\).

\subsection{\workname Core Concept}
\vspace{-0.5em}

\begin{figure}[t]
    \centering
    \includegraphics[width=\linewidth]{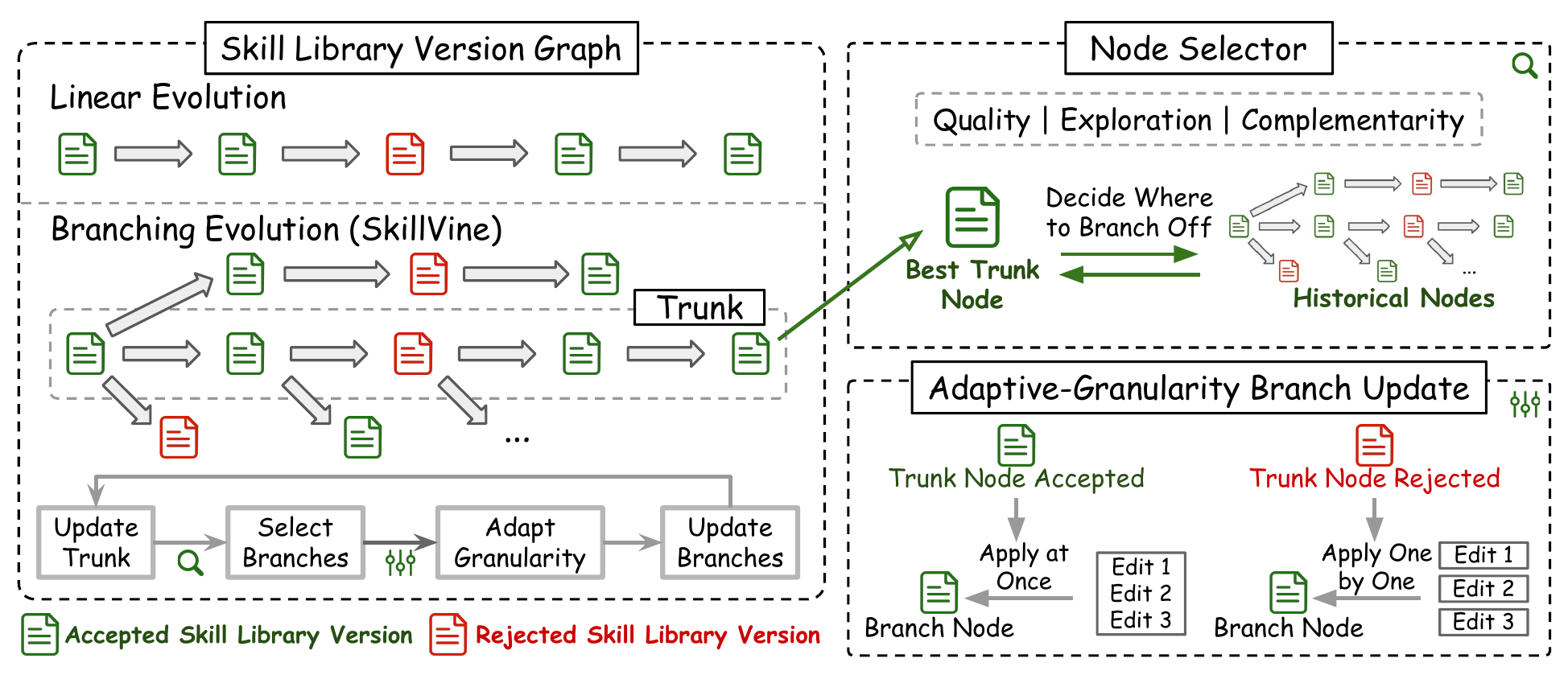}
    \vspace{-1.5em}
    \caption{Overview of \workname.
    \workname organizes skill library versions into a graph, maintains a continuous trunk, selects branch nodes based on quality, exploration, and complementarity, and adapts branch-update granularity based on whether the trunk update is accepted.
    }
    \label{fig:framework_overview}
\end{figure}

As shown in Figure~\ref{fig:framework_overview}, instead of following a single linear patching evolution path, \workname organizes all skill library versions into a version graph and adopts a branching and reversible strategy for skill evolution. The key concept lies in a cooperative trunk-and-branch update as follows:
 
\paragraph{Trunk as a Continuous Batch-Covering Path.}
Training data are fed batch by batch. 
To maintain a continuous optimization path over this stream, \workname keeps a \textbf{trunk} evolution path that is updated on every training batch.
The trunk can see every single batch and accumulates long-horizon information. 
It also provides a reference for selecting branch nodes and deciding how to update them.
This avoids repeatedly branching from early versions, which limits the accumulation of improvements over time.
In contrast, \textbf{branches} grow from selected historical checkpoints. 
Since a branch node is not necessarily selected at every training step, each branch path uses only a subset of the training batches.
Thus, the trunk tracks the cumulative effect of the full batch sequence, while branches explore alternative evolution paths using a subset of training batches.

\paragraph{Trunk-Guided Branch Update.}
This asymmetry between the full-batch trunk and the partial-batch branches motivates a trunk-guided sampling scheme. At each training step, \workname employs a trunk-and-branch sampling scheme. It first updates the trunk incrementally with the current batch. The trunk's accept/reject decision then serves as an information-quality signal for that batch. If the trunk accepts the update, the batch is judged to include more useful information than harmful or malicious information. The node selector can therefore select a promising historical checkpoint and apply a more radical branch update using this batch. 
If the trunk rejects the update, the batch is judged to contain insufficient useful information or potentially harmful information. The corresponding branch update should therefore be made more carefully, e.g., through a more conservative update or a stricter acceptance criterion. 
The node selector identifies promising historical checkpoints for side-branch update, prioritizing versions that remain promising yet underexplored. 
After training, \workname selects the skill library version with the highest validation accuracy from all trunk and branch nodes in the graph as the final output.

\subsection{\workname Methodology}
\vspace{-0.5em}

\paragraph{Skill Library Version Graph Representation.}
To ensure full recoverability during evolution, \workname organizes all skill library versions into a version graph $\mathcal{G}=(\mathcal{V}, \mathcal{E})$, where each node  $v \in \mathcal{V}$ corresponds to a specific skill library version  $S_v$, and each edge  $(u,v) \in \mathcal{E}$ indicates that $S_v$ is derived by updating its parent version $S_u$. This graph preserves every historical version as a recoverable checkpoint, enabling future updates to revert to any earlier node and branch off for alternative exploration paths. 

Each node $v$ in the graph stores three types of metadata. First, for position and role, a \textit{parent pointer} $p(v)$ identifies its parent, and a \textit{node state} $z(v)$ marks it as initial, accepted, rejected, current, or best.
Second, for performance characterization, the node keeps a \textit{validation accuracy} $q(v)$ alongside the explicit sets of \textit{solved and failed validation samples} $(\mathcal{D}_v^{+}, \mathcal{D}_v^{-})$, which collectively offer a more concrete and diagnostic view of its capabilities than a scalar accuracy alone. 
Third, for exploration tracking, the \textit{expansion count} $n_{\mathrm{exp}}(v)$  records how many child versions have already been generated from this node, serving as a key signal for the node selector to prioritize checkpoints that remain promising yet underexplored. 
Together with the saved skill content $S_v$, these metadata allow \workname to restore any historical version and explore new update paths from it.

\paragraph{Trunk-and-Branch Evolution Strategy.}
\workname avoids randomly selecting historical nodes for branching exploration. Instead, it maintains a trunk evolution path for continuous skill refinement while using side branches to explore skill capabilities that the trunk lacks. Training proceeds batch by batch: at each step, a batch of training tasks arrives, and \workname performs three operations: (1) update the trunk, (2) select historical nodes as branching candidates using the node selector, and (3) optimize each selected node, where the optimization granularity is determined by whether the trunk accepts its own update. 
The trunk and each branch collect their own trajectories on the same batch.
Through this evolution strategy, the trunk follows a continuous optimization path that observes every batch and accumulates long-horizon information, whereas branches explore alternative evolution paths that may use only a subset of the training batches.

\paragraph{Trunk Node Update.}
At training step $t$, let $u_t^{\mathrm{tr}}\in\mathcal{V}$ denote the current trunk parent node. Based on the agent interaction trajectories collected at this step, \workname updates $u_t^{\mathrm{tr}}$ and produces a candidate child node $v_t^{\mathrm{tr}}$ for the trunk. It then evaluates the candidate on the validation set and accepts or rejects the update based on the validation result. Specifically, as in SkillOpt-Lite~\citep{shen2026skillopt}, the trunk generates several edits
$\mathcal{P}_t^{\mathrm{tr}}=(\xi_{t,1}^{\mathrm{tr}},\ldots,\xi_{t,m_t}^{\mathrm{tr}})$
from the training trajectories at the current step and then applies all these edits to its parent node. The update is accepted only if the modified skill achieves higher validation accuracy than its parent; otherwise, it is rejected:
\begin{equation}
a_t^{\mathrm{tr}}
=
\mathbb{I}
\left[
q\left(v_t^{\mathrm{tr}}\right)
>
q\left(u_t^{\mathrm{tr}}\right)
\right].
\label{eq:trunk-update-signal}
\end{equation}

\paragraph{Branch Node Selection.}
At each training step, after the trunk update, \workname performs $k$ side-branch updates.
For each branch $i$, if its update was accepted at the previous step, it continues from its latest accepted version. Otherwise, the node selector chooses a historical version as its new parent.
At the first training step, all branch parents are chosen by the node selector.
Let $u_t^{(i)}$ denote the parent of branch $i$ at step $t$. 
Each branch updates its parent using the same training batch as the trunk.
The node selector identifies historical versions with potential for further improvement.
For node $v$, it computes a step-dependent score that combines three terms:
\begin{equation}
\label{eq:node-selection-score}
\mathrm{Select}_t(v)
=
q(v)
+
\lambda_{\mathrm{exp}}
\sqrt{
    \frac{
        \log\left(N_{\mathrm{exp}}+1\right)
    }{
        n_{\mathrm{exp}}(v)+1
    }
}
+
\lambda_{\mathrm{comp}} C_t(v).
\end{equation}
Here, $q(v)$ is the validation accuracy of node $v$. The second term gives a larger bonus to nodes that have produced fewer children, where $n_{\mathrm{exp}}(v)$ is the number of candidate child versions generated and evaluated from node $v$, including both accepted and rejected versions, and $N_{\mathrm{exp}}=\sum_{u\in\mathcal{V}}n_{\mathrm{exp}}(u)$ is the total number of children generated from all nodes. The third term $C_t(v)$ compares node $v$ with the current trunk parent $u_t^{\mathrm{tr}}$. It gives credit when $v$ solves a validation sample that $u_t^{\mathrm{tr}}$ fails, and applies a penalty when $v$ fails a sample that $u_t^{\mathrm{tr}}$ solves:
\begin{equation}
\label{eq:node-complementarity}
C_t(v)
=
\frac{
    \left|
    \mathcal{D}_v^+
    \cap
    \mathcal{D}_{u_t^{\mathrm{tr}}}^-
    \right|
}{
    \left|
    \mathcal{D}_{u_t^{\mathrm{tr}}}^-
    \right|
}
-
\beta
\frac{
    \left|
    \mathcal{D}_{u_t^{\mathrm{tr}}}^+
    \setminus
    \mathcal{D}_v^+
    \right|
}{
    \left|
    \mathcal{D}_{u_t^{\mathrm{tr}}}^+
    \right|
}.
\end{equation}
The coefficients $\lambda_{\mathrm{exp}}$ and $\lambda_{\mathrm{comp}}$
control the weights of exploration and complementarity, while $\beta$ penalizes losing
samples solved by the current trunk.
$C_t(v)$ is essential to our branching strategy. We select nodes based not only on their validation accuracy and expansion counts, but also on whether they solve cases missed by the trunk. 
This allows us to explore a wider range of skill evolution paths within a limited training budget.
In \workname, the node selector coefficients can be set separately for each of the $k$ side-branch updates at a training step.
Larger $\lambda_{\mathrm{exp}}$ and $\lambda_{\mathrm{comp}}$ values favor underexplored, complementary nodes, while smaller values favor validation accuracy.
Empirically, mixing selectors with different selection preferences yields more consistent gains.

\paragraph{Adaptive-Granularity Branch Update (AGBU).}
At each training step, after the trunk is updated and the branch nodes are selected, each branch collects trajectories on the same training batch and generates several skill edits. Some edits generated from the training batch improve validation performance, while others hurt it. Their combined effect determines whether the update is accepted. Since the trunk and branches use the same training data at each step, if this batch produces harmful edits for the trunk, it is also likely to produce harmful edits for the branches. To avoid similar harmful edits in the branches, we use the outcome of the trunk as a signal to adjust the granularity of branch updates. Each branch generates edits
$\mathcal{P}_t^{(i)}=(\xi_{t,1}^{(i)},\ldots,\xi_{t,m_t^{(i)}}^{(i)})$ from the current training batch.

If the trunk update is accepted, i.e., $a_t^{\mathrm{tr}}=1$, the current training batch is likely to produce more beneficial edits. The branches therefore use a coarse-grained update. Each branch applies all its edits to its parent node at once.
Let $v_t^{(i)}$ denote the child node produced by applying all edits to $u_t^{(i)}$:
\begin{equation}
S_{v_t^{(i)}}
=
\mathcal{A}
\left(
S_{u_t^{(i)}},
\mathcal{P}_t^{(i)}
\right),
\qquad
\text{if } a_t^{\mathrm{tr}}=1,
\label{eq:coarse-branch-update}
\end{equation}
where $\mathcal{A}$ denotes applying edits to a skill.
The update is accepted if validation accuracy improves.

If the trunk update is rejected, i.e., $a_t^{\mathrm{tr}}=0$, the training batch is more likely to produce harmful edits. The branches then use a fine-grained update. Each branch applies its edits one at a time and in order. After each edit, the resulting skill is evaluated on the validation set. If the edit is accepted, the resulting version becomes the parent for the next edit. Otherwise, the branch returns to the last accepted version. 
Let $u_{t,j}^{(i)}$ denote the latest accepted node before applying the $j$-th edit and $v_{t,j}^{(i)}$ denote the candidate child node produced by the $j$-th edit , where $u_{t,1}^{(i)}=u_t^{(i)}$:
\begin{equation}
S_{v_{t,j}^{(i)}}
=
\mathcal{A}
\left(
S_{u_{t,j}^{(i)}},
\xi_{t,j}^{(i)}
\right),
\qquad
\text{if } a_t^{\mathrm{tr}}=0.
\label{eq:fine-branch-update}
\end{equation}
All resulting nodes are recorded in the version graph. A coarse-grained update requires only one evaluation and is therefore less costly. A fine-grained update requires multiple evaluations, but it reveals the effect of each edit separately and helps filter out harmful edits. These two update strategies involve a trade-off between evaluation cost and update reliability. We use the outcome of the trunk update as a signal to adapt the branch update strategy at each training step.

\paragraph{ Best Skill Output.}
After training, \workname returns $S^\star=S_{v^\star}$,
where $v^\star=\arg\max_{v\in\mathcal{V}}q(v)$.
Algorithm~\ref{alg:skillgit} summarizes the procedure.

\begin{algorithm}[t]
\caption{\workname}
\label{alg:skillgit}
\begin{algorithmic}[1]

\Require Initial skill library $S_0$; training batches
$\{\mathcal{D}_t\}_{t=1}^{T}$; validation set
$\mathcal{D}_{\mathrm{val}}$; branch budget $k$
\Ensure Final skill library $S^\star$

\State Initialize the skill library version graph
$\mathcal{G}=(\mathcal{V},\mathcal{E})$
with root $v_0$ and skill $S_{v_0}\gets S_0$
\State Evaluate $v_0$ on $\mathcal{D}_{\mathrm{val}}$
and set the trunk parent $u_1^{\mathrm{tr}}\gets v_0$

\For{$t=1,\ldots,T$}

    \State Generate trunk edits $\mathcal{P}_t^{\mathrm{tr}}$
    from trajectories on $\mathcal{D}_t$
    \State Apply all edits to $S_{u_t^{\mathrm{tr}}}$
    and evaluate the candidate $v_t^{\mathrm{tr}}$ on $\mathcal{D}_{val}$

    \State $u_{t+1}^{\mathrm{tr}}\gets v_t^{\mathrm{tr}}$ if $q(v_t^{\mathrm{tr}})>q(u_t^{\mathrm{tr}})$ (accept); otherwise $u_{t+1}^{\mathrm{tr}}\gets u_t^{\mathrm{tr}}$ (reject)

    \State Record $v_t^{\mathrm{tr}}$ and its status in
    $\mathcal{G}$; update its parent link and the parent's
    expansion count

    \For{$i=1,\ldots,k$}
        \If{$t>1$ and branch $i$ produced an accepted update in training step $t-1$}
            \State Set $u_t^{(i)}$ to its latest accepted node from step $t-1$
        \Else
            \State Set $u_t^{(i)}$ to the next parent ranked by~\eqref{eq:node-selection-score}, using $u_t^{\mathrm{tr}}$ as the reference
        \EndIf
    \EndFor

    \ForAll{$i=1,\ldots,k$ \textbf{ in parallel}}

        \State Collect trajectories using $S_{u_t^{(i)}}$
        on $\mathcal{D}_t$ and generate edits $\mathcal{P}_t^{(i)}$

        \If{the trunk update was accepted}
            \State Apply $\mathcal{P}_t^{(i)}$ at once using~\eqref{eq:coarse-branch-update}
            and evaluate $v_t^{(i)}$ on $\mathcal{D}_{val}$
            \State Accept if $q(v_t^{(i)})>q(u_t^{(i)})$;
            otherwise reject
        \Else
            \State Set the initial edit parent
            $u_{t,1}^{(i)}\gets u_t^{(i)}$
            \For{each edit $\xi_{t,j}^{(i)}$ in order}
                \State Apply $\xi_{t,j}^{(i)}$ to
                $S_{u_{t,j}^{(i)}}$ using~\eqref{eq:fine-branch-update}
                and evaluate $v_{t,j}^{(i)}$ on $\mathcal{D}_{val}$
                \State Accept if
                $q(v_{t,j}^{(i)})>q(u_{t,j}^{(i)})$;
                otherwise reject
                \State Set $u_{t,j+1}^{(i)}\gets v_{t,j}^{(i)}$
                if accepted; otherwise set
                $u_{t,j+1}^{(i)}\gets u_{t,j}^{(i)}$
            \EndFor
        \EndIf

    \EndFor

    \State Record all branch nodes and their statuses
    in $\mathcal{G}$; update parent links and expansion counts

\EndFor

\State Select the highest-scoring initial or accepted node
$v^\star$ in $\mathcal{G}$
\State \Return $S_{v^\star}$

\end{algorithmic}
\end{algorithm}

%% file: sec/experiments.tex
\section{Experiments}
\vspace{-0.5em}

In this section, we evaluate \workname against six baselines, assess its components through ablations, and analyze evolution dynamics and system parameters.

\subsection{Experimental Setup}
\vspace{-0.5em}

\textbf{Benchmarks.} We evaluate \workname on five complementary benchmarks: \textbf{SpreadsheetBench}~\citep{ma2024spreadsheetbench} for instruction-based Excel editing; \textbf{LiveMath}~\citep{he2026livemathematicianbench} for multiple-choice mathematical reasoning; \textbf{OfficeQA}~\citep{opsahl2026officeqa} for question answering over office and government documents; \textbf{SearchQA}~\citep{dunn2017searchqa} for open-domain answering with noisy retrieved passages; and \textbf{TABVERSE-QA}~\citep{ahsan2026tabversebenchmarkingcrossformattable} for table-based lookup, comparison, calculation, and verification.

\textbf{Baselines.} We compare \workname with six baselines:
\textbf{Linear:} Skills evolve along a single linear path.
\textbf{Best-of-3}: This baseline maintains a linear evolution path but generates and evaluates three candidate updates per step, selecting the one with the highest validation score.
\textbf{Random}: It uses the same number of branch expansion attempts as \workname but randomly selects parent nodes.
\textbf{Hermes-SE}~\citep{nousresearch2026hermesse}: It uses GEPA~\citep{agrawal2026gepa} to analyze execution traces, generate improved skill variants, and select better variants through evolutionary search. 
\textbf{SkillOpt:}~\citep{yang2026skillopt} It evolves the skill library linearly, compares previous and current skills after each epoch to add protected long-term guidance, and retains optimizer-side experience to guide subsequent edits.
\textbf{SkillOpt-Lite:}~\citep{shen2026skillopt} Built on SkillOpt, it replaces the fixed structured-edit pipeline with a coding agent that directly edits the skill file.

\textbf{Implementation Details.}
We implement \workname on top of SkillOpt-Lite~\citep{shen2026skillopt}.
We use DeepSeek-V4-flash~\citep{xu2026deepseek} provided by Volcano Engine~\citep{volcengine} and Qwen3.8-Flash~\citep{qwen38flash} provided by Alibaba Cloud Model Studio~\citep{alibaba_model_studio} as base models. We disable thinking mode for Qwen3.8-Flash.
We follow the official SkillOpt-Lite repository’s epoch, batch-size, and training-step settings.
Unless otherwise specified, all methods with branch exploration use two branch update attempts per training step.
For \workname, we use a mix of node selectors that emphasize exploration and complementarity and those that emphasize higher validation accuracy. 
We refer to them as \textbf{discovery selectors} and \textbf{refinement selectors}, respectively. We split the $k$ side-branch update attempts per training step equally between the two, giving the discovery selectors one extra attempt when $k$ is odd.
We set $\lambda_{\mathrm{exp}}=0.08$ and $\lambda_{\mathrm{comp}}=0.5$ for discovery selectors, and set both to zero for refinement selectors. We set $\beta=1.6$.

\subsection{Main Results}
\vspace{-0.5em}

\begin{table*}[t]
\centering
\caption{Main results across benchmarks.
Higher values indicate better performance.}
\label{tab:main_results}
\footnotesize
\setlength{\tabcolsep}{3pt}
\renewcommand{\arraystretch}{1.08}
\begin{tabularx}{\textwidth}{
c V
c V
*{5}{Y}
}
\toprule
\multirow[c]{2}{*}{\textbf{Model}}
&
\multirow[c]{2}{*}{\textbf{Method}}
&
\multicolumn{5}{c}{\textbf{Benchmark Score}}
\\
\cmidrule(lr){3-7}
&
&
\textbf{Spreadsheet}
&
\textbf{LiveMath}
&
\textbf{SearchQA}
&
\textbf{OfficeQA}
&
\textbf{TABVERSE}
\\
\midrule
\multirow{7}{*}{
    \makecell[l]{DeepSeek-V4\\-flash}
}
& Linear
& 48.04 & 43.40 & 73.36 & 47.30 & 63.88 \\
& Best-of-3
& 64.41 & 50.00 & 73.14 & 45.27 & 64.49 \\
& Random
& 53.38 & 46.23 & 74.14 & 52.70 & 64.08 \\
& SkillOpt
& 62.99 & 47.17 & 73.50 & 46.62 & 63.47 \\
& SkillOpt-Lite
& 66.19 & 50.48 & 73.14 & 37.84 & 61.84 \\
& Hermes-SE
& 50.53 & 51.89 & 73.57 & 53.38 & 60.82 \\
\rowcolor{bestrow}
\cellcolor{white}\strut
& \textbf{\workname}
& \bfseries 71.53
& \bfseries 53.77
& \bfseries 74.43
& \bfseries 55.41
& \bfseries 67.35
\\
\midrule
\multirow{7}{*}{
    \makecell[l]{Qwen3.8\\-Flash}
}
& Linear
& 66.90 & 36.79 & 72.14 & 50.00 & 60.82 \\
& Best-of-3
& 65.12 & 32.08 & 72.21 & 47.30 & 60.82 \\
& Random
& 66.90 & 37.74 & 72.14 & 50.00 & 61.22 \\
& SkillOpt
& 65.12 & 36.79 & 72.21 & 43.24 & 61.63 \\
& SkillOpt-Lite
& 65.84 & 33.02 & 71.93 & 51.91 & 59.39 \\
& Hermes-SE
& 66.19 & \textbf{42.45} & 71.86 & 47.97 & 62.24 \\
\rowcolor{bestrow}
\cellcolor{white}\strut
& \textbf{\workname}
& \textbf{67.97} & 39.62 & \textbf{72.40} & \textbf{54.05} & \textbf{63.67} \\
\bottomrule
\end{tabularx}
\end{table*}

Table~\ref{tab:main_results} presents the test performance of \workname and 6 baselines across 5 benchmarks and 2 base LLMs.
To reduce the effect of randomness in agent rollouts and save API calls, we first run Linear for each benchmark-model combination.
Best-of-3, Random, and \workname reuse Linear's trunk results.
Nodes from the trunk become available only when the corresponding step is reached, so no method can access results from future steps.
Note that Best-of-3 reuses the Linear trunk node only when neither additional attempt achieves a higher validation score. Otherwise, it generates subsequent results independently.
SkillOpt and SkillOpt-Lite reuse Linear's first-epoch results.
These reuse does not change the methods.
Table~\ref{tab:main_results} shows that \workname achieves the highest test score on all the five benchmarks with DeepSeek-V4-flash and four of the five benchmarks with Qwen3.8-Flash.
\workname achieves large gains over Linear, especially on SpreadsheetBench with DeepSeek-V4-flash, where the test score rises from 48.04 to 71.53, an increase of 23.49.
This indicates the limitations of linear evolution and shows that branching exploration can find better skill library versions that linear evolution misses.
\workname outperforms Best-of-3 in all 10 benchmark--model combinations. 
This suggests that \workname is more effective than simply increasing the number of update attempts.
\workname also outperforms Random in all 10 benchmark--model combinations.
This shows that \workname's guided branch exploration is more effective than branching from randomly selected nodes.
The gains vary across benchmarks and base LLMs.
Gains are larger on OfficeQA and SpreadsheetBench with DeepSeek-V4-flash, where errors are concentrated in a few types and the weaker trunk leaves more room for improvement.
Gains are smaller on SearchQA, where errors are more diverse and the trunk has achieved high performance.
\workname achieves the highest scores among all methods in most settings, supporting its broad applicability.
The validation scores of the final best nodes are reported in Table~\ref{tab:validation_results} in the appendix.

\subsection{Ablation Study}
\vspace{-0.5em}

\begin{table*}[t]
\centering
\caption{Ablation study results.
Higher values indicate better performance.}
\label{tab:ablation}
\footnotesize
\setlength{\tabcolsep}{3pt}
\renewcommand{\arraystretch}{1.08}
\begin{tabularx}{\textwidth}{
c V
c V
*{5}{Y}
}
\toprule
\multirow[c]{2}{*}{\textbf{Model}}
&
\multirow[c]{2}{*}{\textbf{Ablation}}
&
\multicolumn{5}{c}{\textbf{Benchmark Score}}
\\
\cmidrule(lr){3-7}
&
&
\textbf{Spreadsheet}
&
\textbf{LiveMath}
&
\textbf{SearchQA}
&
\textbf{OfficeQA}
&
\textbf{TABVERSE}
\\
\midrule
& w/o Selector \& AGBU
& 53.38 & 46.23 & 74.14 & 52.70 & 64.08 \\
& w/o Selector
& 67.26 & 46.23 & 72.93 & 47.30 & 63.47 \\
& w/o AGBU
& 65.84 & 50.00 & \textbf{74.57} & 52.03 & 63.27 \\
& w/o Comp
& \textbf{71.53} & \textbf{53.77} & 73.36 & \textbf{55.41} & 60.82 \\
\rowcolor{bestrow}
\cellcolor{white}
\multirow{-5}{*}{\makecell[l]{DeepSeek-V4\\-flash}}
& \textbf{\workname}
& \textbf{71.53}
& \textbf{53.77}
& 74.43
& \textbf{55.41}
& \textbf{67.35}
\\
\midrule
& w/o Selector \& AGBU
& 66.90 & 37.74 & 72.14 & 50.00 & 61.22 \\
& w/o Selector
& 65.84 & 35.85 & 72.14 & 50.00 & 63.27 \\
& w/o AGBU
& 65.48 & 37.74 & \textbf{72.86} & 51.35 & 62.65 \\
& w/o Comp
& 66.55 & \textbf{39.62} & 71.07 & \textbf{54.05} & 61.43 \\
\rowcolor{bestrow}
\cellcolor{white}
\multirow{-5}{*}{\makecell[l]{Qwen3.8\\-Flash}}
& \textbf{\workname}
& \textbf{67.97}
& \textbf{39.62}
& 72.40
& \textbf{54.05}
& \textbf{63.67}
\\
\bottomrule
\end{tabularx}
\end{table*}

Table~\ref{tab:ablation} presents the ablation results for \workname.
For the w/o Selector setting, we replace our node selector with random selection while retaining AGBU.
For both base LLMs, \workname outperforms most ablation variants with components removed.
On SearchQA, as the test scores are close and agent rollouts are stochastic, \workname has a slightly lower test score than w/o AGBU. 
However, the validation score achieved by the final best node of \workname is still the highest.
Removing complementarity changes parent-node selection in five of the ten model–benchmark combinations. To reduce rollout variance, we fully reuse \workname's training results for the five unchanged combinations; for the others, we reuse results up to the first selection divergence and rerun from that step onward with the complementarity term removed.
Table~\ref{tab:ablation} suggests that complementarity helps discover better alternative paths.
Figure~\ref{fig:ablation_rollout_budget} compares training cost, measured by the number of agent rollouts, and performance across the five settings.
Branch exploration and AGBU require additional rollouts, but both improve performance. The w/o Selector setting uses slightly more rollouts than \workname but performs worse, showing that our selector uses the branch budget more effectively.

\subsection{Analysis}
\vspace{-0.5em}

\subsubsection{Skill Evolution Dynamics}
\vspace{-0.5em}

\begin{figure*}[t]
\centering

\begin{minipage}[t]{0.32\textwidth}
    \vspace{0pt}
    \centering
    \includegraphics[width=\linewidth]{
        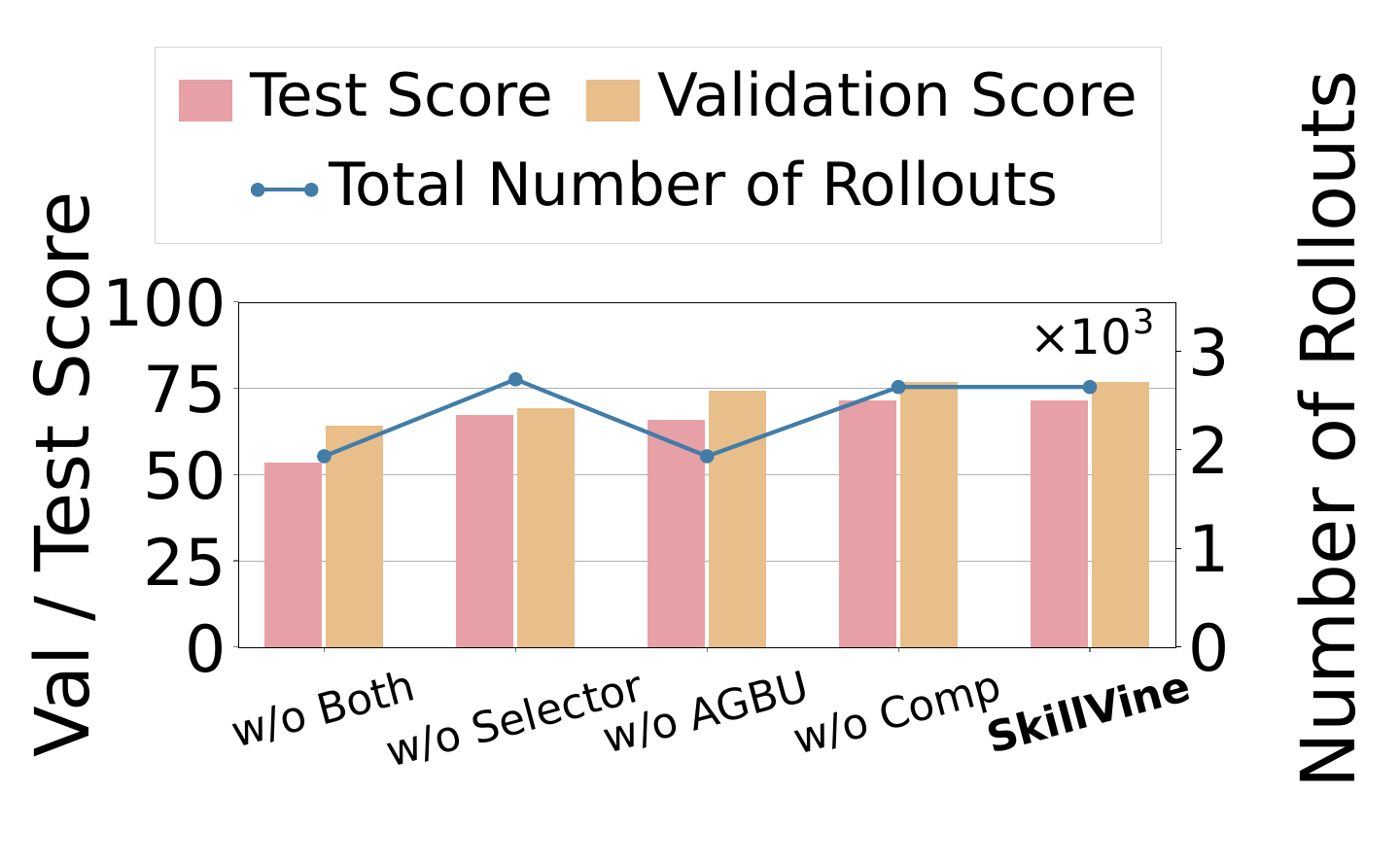
    }
    \caption{
        Training cost and performance on SpreadsheetBench with two branch update attempts per training step.
    }
    \label{fig:ablation_rollout_budget}
\end{minipage}%
\hfill
\begin{minipage}[t]{0.32\textwidth}
    \vspace{0pt}
    \centering
    \includegraphics[width=\linewidth]{
        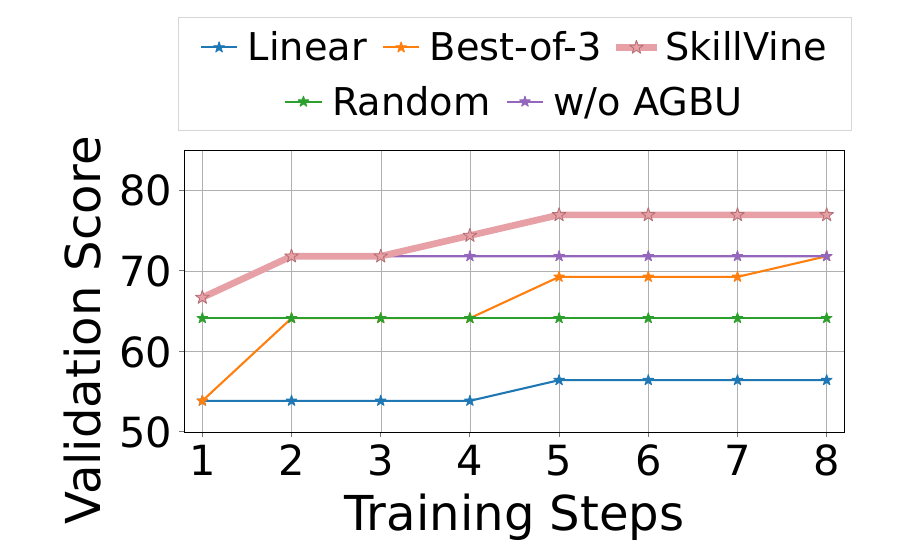
    }
    \caption{
        Validation score of the best node found up to each training step on SpreadsheetBench.
    }
    \label{fig:best_node_validation}
\end{minipage}%
\hfill
\begin{minipage}[t]{0.32\textwidth}
    \vspace{0pt}
    \centering
    \includegraphics[width=\linewidth]{
        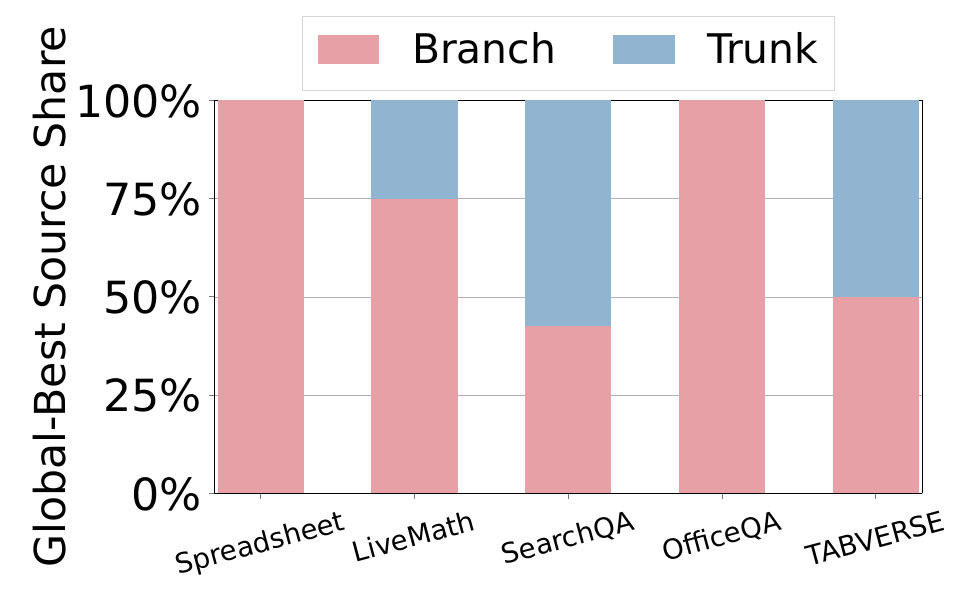
    }
    \caption{
        Proportion of training steps in which the best node comes from the trunk or side branches.
    }
    \label{fig:global_best_sources}
\end{minipage}

\end{figure*}

Figure~\ref{fig:best_node_validation} shows the best validation score found up to each training step on SpreadsheetBench.
\workname achieves the highest score at every step across all eight training steps, reaching 76.92\%. 
The final best node achieves the highest test score of 71.53\%.
We further examine whether the best nodes come from the trunk or branches.
Figure~\ref{fig:global_best_sources} shows the percentage of training steps in which the global-best node comes from each source.
Branches provide the global-best node in a large proportion of training steps.
On SpreadsheetBench and OfficeQA, the global-best node even come from branch exploration at every training step.
A closer analysis shows that, across all ten benchmark--model combinations, the final best nodes are reached through branch paths.
These results indicate that branch exploration helps find better skill library versions than those available along the trunk.
Among the ten paths, two paths use only discovery selectors, two only refinement selectors, and six both, demonstrating that both selector types contribute to finding the best paths.
AGBU rejects at least one harmful edit in eight paths, helping preserve promising evolution paths.
Table~\ref{tab:evolution_paths} details all ten paths.

\subsubsection{Impact of System Parameters}
\vspace{-0.5em}

\begin{figure*}[t]
\centering
\makebox[\textwidth][c]{%
\subcaptionbox{
    Impact of branches numbers.
    \label{fig:number_of_branches}
}[0.355\textwidth]{%
    \includegraphics[width=\linewidth]{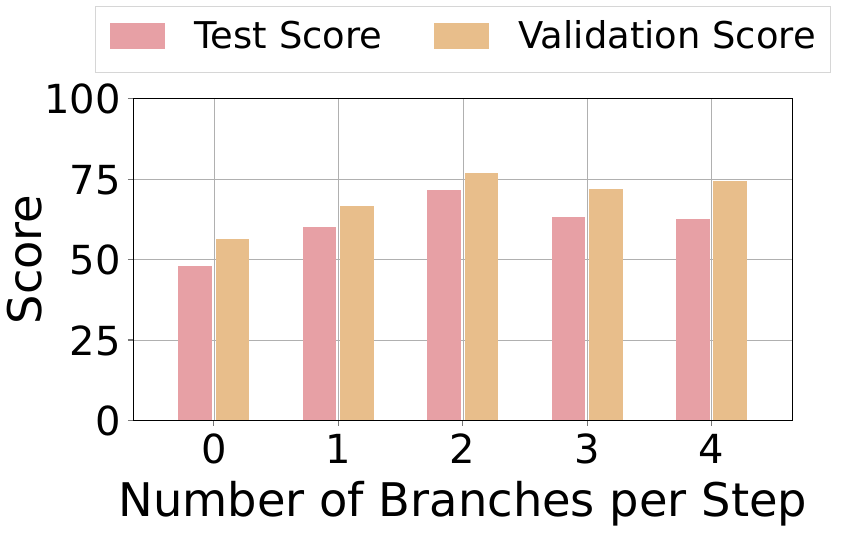}
}%
\hspace{0.000\textwidth}%
\subcaptionbox{
    Impact of branch update interval.
    \label{fig:branch_update_interval}
}[0.355\textwidth]{%
    \includegraphics[width=\linewidth]{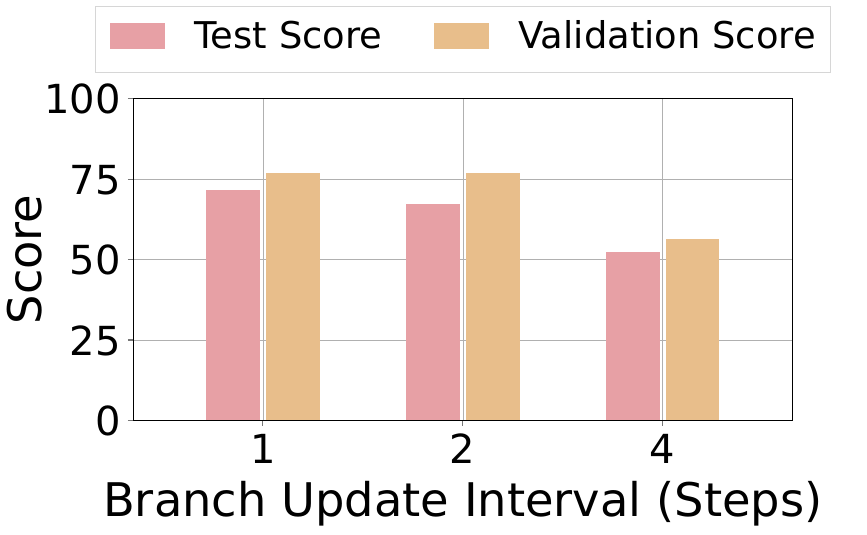}
}%
\hspace{0.000\textwidth}%
\subcaptionbox{
    Impact of training steps.
    \label{fig:training_steps}
}[0.355\textwidth]{%
    \includegraphics[width=\linewidth]{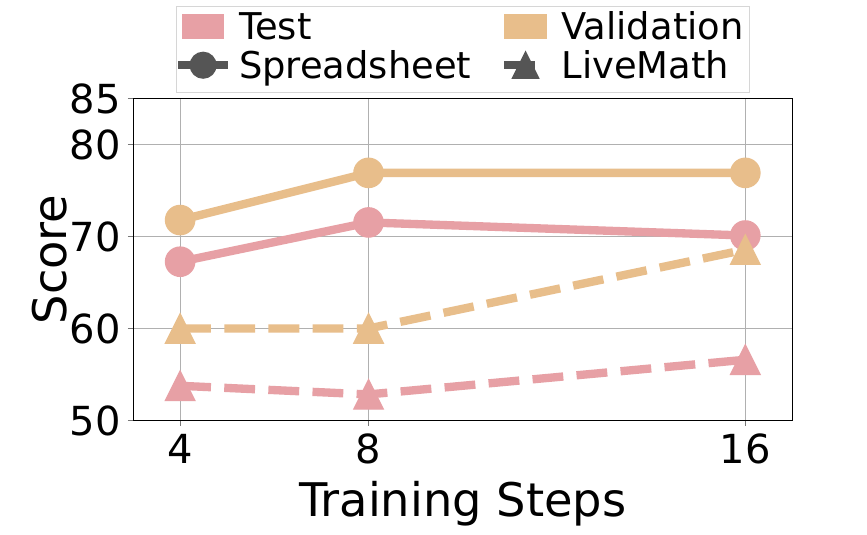}
}%
}
\caption{Impact of system parameters on validation and test scores with DeepSeek-V4-flash. Panels (a) and (b) use SpreadsheetBench; panel (c) includes both SpreadsheetBench and LiveMath.}
\label{fig:system_parameters}
\end{figure*}

Figure~\ref{fig:system_parameters} shows how three system parameters affect validation and test scores: the number of branch update attempts per training step, the branch update interval, and the number of training steps.
Figure~\ref{fig:number_of_branches} shows that performance generally improves at first and then levels off as the number of branch update attempts per training step increases. Even one attempt improves performance, while two attempts already provide strong gains.
Figure~\ref{fig:branch_update_interval} shows that shorter branch update intervals lead to better performance. A shorter interval allows branch updates at more training steps and produces more candidate nodes, increasing the chance of finding a high-performing branch node.
Figure~\ref{fig:training_steps} shows that using smaller batches and more training steps generally improves performance. With a fixed number of branch update attempts per step, more steps provide more rounds of branch exploration and produce more candidate nodes, increasing the chance of finding a high-performing node. More steps also divide evolution into more stages, allowing updates to be added and evaluated gradually and reducing the risk of harmful edits accumulating in a single update.

%% file: sec/conclusion.tex
\section{Conclusion}
\vspace{-1em}

In this paper, we introduce \workname, an automatic skill-evolution framework that enables parallel search through branching exploration. Its core components include a trunk-branch collaborative search mechanism, a complementarity-aware node selection method, and an adaptive-granularity update rule. 
Overall, \workname outperforms six baselines in nine of ten model–benchmark combinations and finds better skill-library versions through branching than along the linear trunk, offering a new approach to agent self-evolution.

%% file: sec/ai_statement.tex
\section*{AI Use Statement}
\vspace{-0.5em}

We used generative AI tools to assist with writing and language polishing, literature retrieval and discovery, research ideation and execution, and drafting sections of the paper. We did not use generative AI tools to generate synthetic datasets or prove mathematical claims. We reviewed all AI-assisted work and take responsibility for the final content of this paper, including text, claims, and artifacts produced with the aid of generative AI.

%% file: sec/ref.tex
{\small
\bibliographystyle{iclr2027_conference}
\bibliography{ref}
}

%% file: sec/appendix.tex
\appendix
\section{Appendix}

\subsection{\workname Evolution Paths across 10 Benchmark--Model Combinations}

\begin{table}[H]
\centering
\caption{
Evolution paths to the final best nodes across the ten benchmark--model combinations.
Each vXXXX is a skill library version number within that run.
\MainNode{vXXXX} denotes a node on the trunk;
\DiscoveryNode{vXXXX} denotes a side-branch node generated using a discovery selector;
\RefinementNode{vXXXX} denotes a side-branch node generated using a refinement selector.
\(\dagger\) means that AGBU rejects at least one edit during the update.
}
\label{tab:evolution_paths}

\vspace{2pt}
\footnotesize
\setlength{\tabcolsep}{5pt}
\renewcommand{\arraystretch}{1.18}

\begin{tabularx}{\textwidth}{C{2.4cm} V C{3.0cm} V X}
\toprule

\textbf{Model}
&
\textbf{Benchmark}
&
\textbf{Evolution Path}
\\

\midrule

\multirow[c]{5}{*}{%
  \makecell[c]{DeepSeek-V4\\-flash}%
}
&
SpreadsheetBench
&
\MainNode{v0001}
\PathArrow
\DiscoveryNode{v0004}
\PathArrow
\AGBUNode{\RefinementNode{v0007}}
\PathArrow
\AGBUNode{\DiscoveryNode{v0026}}
\PathArrow
\DiscoveryNode{v0030}
\\

\cmidrule(lr){2-3}

&
LiveMath
&
\MainNode{v0001}
\PathArrow
\DiscoveryNode{v0004}
\PathArrow
\RefinementNode{v0006}
\PathArrow
\AGBUNode{\DiscoveryNode{v0011}}
\\

\cmidrule(lr){2-3}

&
SearchQA
&
\MainNode{v0001}
\PathArrow
\DiscoveryNode{v0004}
\PathArrow
\DiscoveryNode{v0010}
\PathArrow
\AGBUNode{\DiscoveryNode{v0041}}
\PathArrow
\AGBUNode{\RefinementNode{v0187}}
\\

\cmidrule(lr){2-3}

&
OfficeQA
&
\MainNode{v0001}
\PathArrow
\RefinementNode{v0009}
\\

\cmidrule(lr){2-3}

&
TABVERSE
&
\MainNode{v0001}
\PathArrow
\DiscoveryNode{v0004}
\PathArrow
\AGBUNode{\DiscoveryNode{v0010}}
\\

\midrule

\multirow[c]{5}{*}[-1.2\baselineskip]{%
  \makecell[c]{Qwen3.8\\-Flash}%
}
&
SpreadsheetBench
&
\MainNode{v0001}
\PathArrow
\RefinementNode{v0003}
\PathArrow
\AGBUNode{\RefinementNode{v0013}}
\PathArrow
\AGBUNode{\DiscoveryNode{v0027}}
\\

\cmidrule(lr){2-3}

&
LiveMath
&
\MainNode{v0001}
\PathArrow
\RefinementNode{v0003}
\\

\cmidrule(lr){2-3}

&
SearchQA
&
\MainNode{v0001}
\PathArrow
\MainNode{v0002}
\PathArrow
\MainNode{v0005}
\PathArrow
\MainNode{v0017}
\PathArrow
\MainNode{v0020}
\PathArrow
\MainNode{v0113}
\PathArrow
\AGBUNode{\DiscoveryNode{v0227}}
\\

\cmidrule(lr){2-3}

&
OfficeQA
&
\MainNode{v0001}
\PathArrow
\DiscoveryNode{v0004}
\PathArrow
\RefinementNode{v0010}
\PathArrow
\AGBUNode{\RefinementNode{v0012}}
\PathArrow
\AGBUNode{\RefinementNode{v0018}}
\PathArrow
\AGBUNode{\RefinementNode{v0024}}
\\

\cmidrule(lr){2-3}

&
TABVERSE
&
\MainNode{v0001}
\PathArrow
\AGBUNode{\DiscoveryNode{v0007}}
\PathArrow
\AGBUNode{\DiscoveryNode{v0017}}
\PathArrow
\AGBUNode{\RefinementNode{v0029}}
\PathArrow
\AGBUNode{\RefinementNode{v0037}}
\PathArrow
\AGBUNode{\DiscoveryNode{v0065}}
\\

\bottomrule
\end{tabularx}
\end{table}

Table~\ref{tab:evolution_paths} presents the evolution paths leading to the final best nodes across the ten benchmark--model combinations. 
Each path traces the sequence of skill library version numbers from the initial node to the final best node.
As shown in Table~\ref{tab:evolution_paths}, all ten best nodes are reached through paths that diverge from the trunk, showing that better paths are common but often missed by linear evolution, whereas \workname can discover these better paths through branch exploration.
Among these ten paths, two paths contain only discovery selector updates, two paths contain only refinement selector updates, and the remaining six paths contain both, showing that both selectors contribute to finding paths to the best nodes.
In eight of these ten paths, AGBU rejects at least one harmful edit, showing that it helps prevent harmful edits from disrupting promising paths to the best nodes.

\subsection{Examples of Complementary Nodes}
\begin{table}[t]
\centering
\caption{
Two examples of selected complementary nodes.
}
\label{tab:complementary_node_examples}

\begingroup
\footnotesize
\setlength{\tabcolsep}{4pt}
\setlength{\arrayrulewidth}{0.4pt}
\renewcommand{\arraystretch}{1.18}
\renewcommand{\tabularxcolumn}[1]{m{#1}}

\definecolor{CompTrunk}{HTML}{0072B2}
\definecolor{CompRefinement}{HTML}{009E73}
\definecolor{CompDiscovery}{HTML}{D55E00}

\newcommand{\CompT}{\textcolor{CompTrunk}{\textbf{T}}}
\newcommand{\CompBR}{\textcolor{CompRefinement}{\textbf{B}}}
\newcommand{\CompBD}{\textcolor{CompDiscovery}{\textbf{B}}}

\begin{tabularx}{\linewidth}{
>{\centering\arraybackslash}m{0.12\linewidth}|
>{\raggedright\arraybackslash}X|
>{\centering\arraybackslash}m{0.085\linewidth}|
>{\raggedright\arraybackslash}m{0.28\linewidth}|
>{\raggedright\arraybackslash}m{0.16\linewidth}
}
\toprule

\multicolumn{5}{c}{
\textbf{Case 1: DeepSeek-V4-flash $\times$ SpreadsheetBench}
}
\tabularnewline
\midrule

\multicolumn{1}{c|}{
  \begin{tabular}[c]{@{}c@{}}
    \textbf{Version Node}
  \end{tabular}
}
&
\multicolumn{1}{c|}{
  \begin{tabular}[c]{@{}c@{}}
    \textbf{Skill Capability}
  \end{tabular}
}
&
\multicolumn{1}{c|}{
  \begin{tabular}[c]{@{}c@{}}
    \textbf{Unique / Lost}
  \end{tabular}
}
&
\multicolumn{1}{c|}{
  \begin{tabular}[c]{@{}c@{}}
    \textbf{Typical Validation Samples}
  \end{tabular}
}
&
\multicolumn{1}{c}{
  \begin{tabular}[c]{@{}c@{}}
    \textbf{Final Gain}
  \end{tabular}
}
\tabularnewline
\midrule

Trunk\newline
\MainNode{v0002}
\par\medskip
Branch\newline
\RefinementNode{v0007}
&
\textbf{\RefinementNode{v0007} adds Python-based calculation beyond
\MainNode{v0002}'s workbook inspection and preservation.}
\par\smallskip
\MainNode{v0002} rules:\newline
Inspect headers and cell types; read cached formula
values while preserving the workbook.
\par\smallskip
\RefinementNode{v0007} rules:\newline
Compute in Python and write values directly;
infer the intended operation from the actual table
when the supplied formula is flawed.
&
\multicolumn{1}{c|}{
  \begin{tabular}[c]{@{}c@{}}
    \MainNode{v0002}\\
    0 / 7\\[8pt]
    \RefinementNode{v0007}\\
    7 / 0
  \end{tabular}
}
&
\textbf{10747: Two-key profit lookup}\newline
\MainNode{v0002}: \texttt{None} (fail)\newline
\RefinementNode{v0007}: \texttt{-10600} (pass)
\par\smallskip
\textbf{51556: Conditional average}\newline
\MainNode{v0002}: \texttt{None} (fail)\newline
\RefinementNode{v0007}: \texttt{82} (pass)
\par\smallskip
The branch also solves all 21 examples solved by the trunk.
&
Final best \DiscoveryNode{v0030} outperforms the best trunk \MainNode{v0028} by 20.51 on validation and 23.49 on test.
\tabularnewline
\midrule

\multicolumn{5}{c}{
\textbf{Case 2: DeepSeek-V4-flash $\times$ SearchQA}
}
\tabularnewline
\midrule

\multicolumn{1}{c|}{
  \begin{tabular}[c]{@{}c@{}}
    \textbf{Version Node}
  \end{tabular}
}
&
\multicolumn{1}{c|}{
  \begin{tabular}[c]{@{}c@{}}
    \textbf{Skill Capability}
  \end{tabular}
}
&
\multicolumn{1}{c|}{
  \begin{tabular}[c]{@{}c@{}}
    \textbf{Unique / Lost}
  \end{tabular}
}
&
\multicolumn{1}{c|}{
  \begin{tabular}[c]{@{}c@{}}
    \textbf{Typical Validation Samples}
  \end{tabular}
}
&
\multicolumn{1}{c}{
  \begin{tabular}[c]{@{}c@{}}
    \textbf{Final Gain}
  \end{tabular}
}
\tabularnewline
\midrule

Trunk\newline
\MainNode{v0031}
\par\medskip
Branch\newline
\DiscoveryNode{v0010}
&
\textbf{\DiscoveryNode{v0010} handles wordplay and answer forms
missed by \MainNode{v0031}; \MainNode{v0031} better resolves associated entities in the examples shown.}
\par\smallskip
\MainNode{v0031} rule:\newline
For a bare entity name, identify the associated
category or relationship from context.
\par\smallskip
\DiscoveryNode{v0010} rules:\newline
Match the source answer form; for word
transformations, return both the original
and transformed terms.
&
\multicolumn{1}{c|}{
  \begin{tabular}[c]{@{}c@{}}
    \MainNode{v0031}\\
    5 / 9\\[8pt]
    \DiscoveryNode{v0010}\\
    9 / 5
  \end{tabular}
}
&
\textbf{``Adam Levine''}\newline
\MainNode{v0031}: Maroon 5 (pass)\newline
\DiscoveryNode{v0010}: Adam Levine (fail)
\par\smallskip
\textbf{Holes on postage stamps}\newline
\MainNode{v0031}: perforation (fail)\newline
\DiscoveryNode{v0010}: perforations (pass)
\par\smallskip
\textbf{Hawaiian wreath becoming a sheltered area}\newline
\MainNode{v0031}: lee (fail)\newline
\DiscoveryNode{v0010}: lei \& lee (pass)
&
Final best \RefinementNode{v0187} outperforms
the best trunk \MainNode{v0134} by
5.00 on validation and
1.07 on test.
\tabularnewline
\bottomrule

\end{tabularx}
\endgroup
\end{table}

Table~\ref{tab:complementary_node_examples} presents two examples of complementary nodes selected by \workname.
In Case 1, R:v0007 solves seven validation samples missed by M:v0002 while retaining all 21 samples solved by the trunk.
In Case 2, D:v0010 solves nine samples missed by M:v0031 but loses
five samples solved by the trunk.
In both cases, evolution from the selected branch produces a final node that outperforms the best trunk node on both validation and test sets.
These examples show that the complementarity-aware selector preserves useful alternative capabilities and opens better evolution paths.

\subsection{Examples of Harmful Edits Rejected by AGBU}
\begin{table}[t]
\centering
\caption{
Two examples of harmful edits rejected by AGBU.
}
\label{tab:agbu_examples}

\begingroup
\footnotesize
\setlength{\tabcolsep}{4pt}
\setlength{\arrayrulewidth}{0.4pt}
\renewcommand{\arraystretch}{1.18}
\renewcommand{\tabularxcolumn}[1]{m{#1}}

\newcommand{\AGBUCaseHeader}{
\multicolumn{1}{c|}{
\begin{tabular}[c]{@{}c@{}}
\textbf{Version Node}
\end{tabular}
}
&
\multicolumn{1}{c|}{
\begin{tabular}[c]{@{}c@{}}
\textbf{Rejected Trunk Update}
\end{tabular}
}
&
\multicolumn{1}{c|}{
\begin{tabular}[c]{@{}c@{}}
\textbf{AGBU-Filtered Branch Update}
\end{tabular}
}
&
\multicolumn{1}{c}{
\begin{tabular}[c]{@{}c@{}}
\textbf{Accepted Gain}
\end{tabular}
}
\tabularnewline
}

\begin{tabularx}{\linewidth}{
>{\centering\arraybackslash}m{0.145\linewidth}|
>{\raggedright\arraybackslash}m{0.29\linewidth}|
>{\raggedright\arraybackslash}X|
>{\raggedright\arraybackslash}m{0.16\linewidth}
}
\toprule

\multicolumn{4}{c}{
\textbf{Case 1: DeepSeek-V4-flash $\times$ TABVERSE}
}
\tabularnewline
\midrule
\AGBUCaseHeader
\midrule

Mainline\newline
\MainNode{v0002}
$\rightarrow$
\MainNode{v0005}\newline
(rejected)
\par\medskip
Branch\newline
\DiscoveryNode{v0004}
$\rightarrow$
\AGBUNode{\RefinementNode{v0007}}
&
\textbf{Proposed rule:}
Return the full cell text, such as
``Middelkerke, Belgium'', rather than one part.
\par\smallskip
\textbf{Result:}
The update included this rule.
It lost three previously solved samples and only gained two:
45/70 $\rightarrow$ 44/70
(\textbf{$-1.43$}), so it was rejected.
&
AGBU \textbf{kept one edit and rejected two}.
\par\smallskip
The kept edit selects the correct column and computes
durations from date ranges:
47/70 $\rightarrow$ 49/70.
\par\smallskip
An edit with the same full-cell rule then reduced
the score to 43/70
(\textbf{$-8.57$}), so AGBU rejected it.
&
The accepted child
\AGBUNode{\RefinementNode{v0007}}
improved over
\DiscoveryNode{v0004}:
47/70 $\rightarrow$ 49/70,
\textbf{+2 samples}
(\textbf{+2.86}).
\tabularnewline
\midrule

\multicolumn{4}{c}{
\textbf{Case 2: Qwen3.8-Flash $\times$ SearchQA}
}
\tabularnewline
\midrule
\AGBUCaseHeader
\midrule

Mainline\newline
\MainNode{v0020}
$\rightarrow$
\MainNode{v0058}\newline
(rejected)
\par\medskip
Branch\newline
\MainNode{v0002}
$\rightarrow$
\AGBUNode{\DiscoveryNode{v0066}}
&
\textbf{Proposed rule:}
Shorten names: use ``Webster'' instead of
``Noah Webster'' and ``the Colorado'' instead of
``Colorado River''.
\par\smallskip
\textbf{Result:}
The update included this rule.
The score fell from 152/200 to 148/200
(\textbf{$-2.00$}), so it was rejected.
&
AGBU \textbf{kept two edits and rejected two}.
\par\smallskip
An edit with the same name-shortening rule reduced
the score from 147/200 to 141/200
(\textbf{$-3.00$}), so it was rejected.
\par\smallskip
The kept edits select the specific answer type and
use ``\&'' for paired answers such as
``Nina \& Pinta''.
Together, they raised the score to 151/200.
&
The accepted child
\AGBUNode{\DiscoveryNode{v0066}}
improved over
\MainNode{v0002}:
147/200 $\rightarrow$ 151/200,
\textbf{+4 samples}
(\textbf{+2.00}).
\tabularnewline
\bottomrule

\end{tabularx}
\endgroup
\end{table}

Table~\ref{tab:agbu_examples} presents two examples of how AGBU filters harmful edits from rejected trunk-update batches.
In Case 1, the trunk update introduces an overgeneralized rule that always returns the full cell text.
This rule loses more validation samples than it gains, causing the trunk update to be rejected.
During the branch update, AGBU rejects the same harmful edit but retains a useful edit for selecting the correct column and computing durations, allowing R:v0007 to improve from 47/70 to 49/70 and be accepted.
In Case 2, the trunk update introduces an overly broad name-shortening rule that reduces the validation score from 152/200 to 148/200.
AGBU rejects the same rule on the branch while retaining two useful edits for selecting the answer type and formatting paired answers, allowing D:v0066 to improve from 147/200 to 151/200 and be accepted.
These cases show that AGBU can separate harmful edits from useful ones and preserve the beneficial parts of an otherwise rejected update.

\subsection{An Example of Evolution Process}
\begin{figure}[t]
    \centering
    \includegraphics[width=\linewidth]{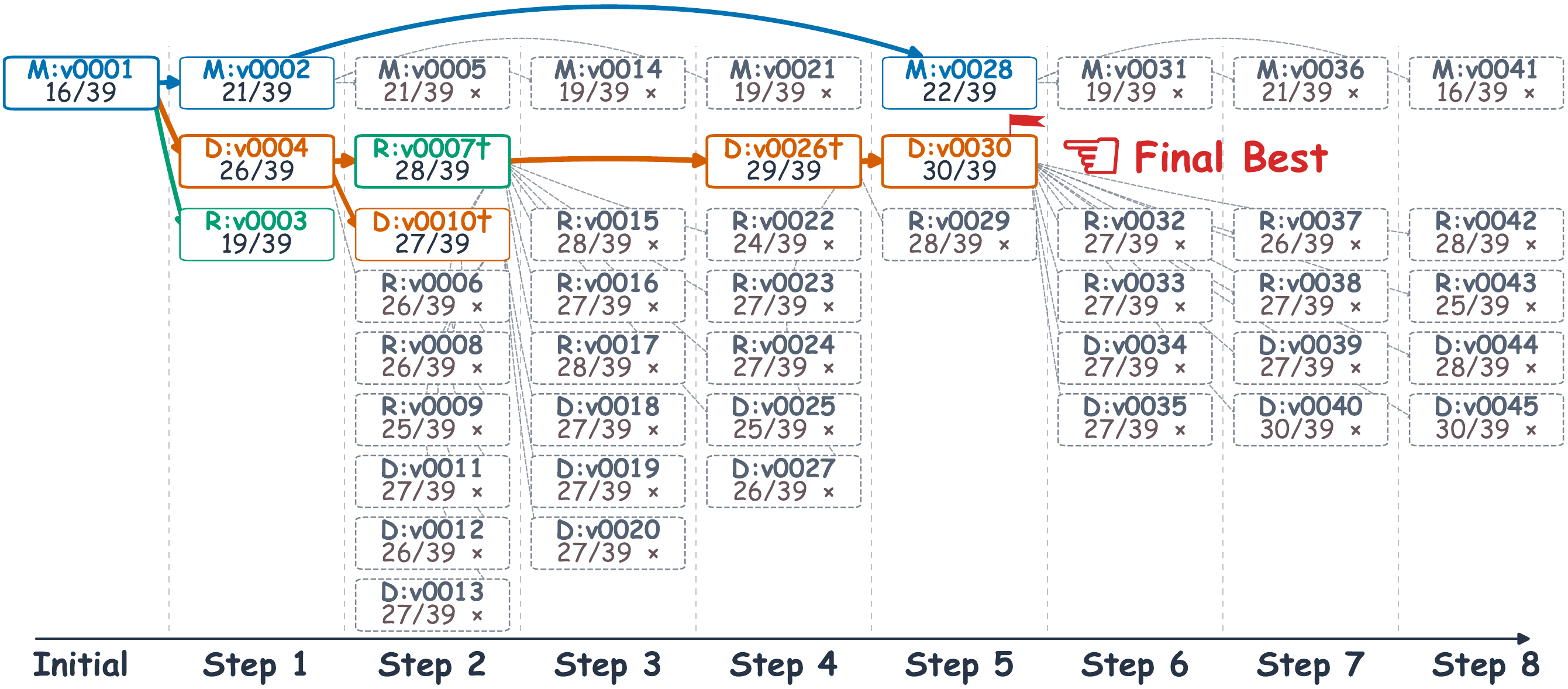}
    \caption{
        Evolution path of DeepSeek-V4-flash on SpreadsheetBench.
        Nodes are labeled with their version numbers and validation scores.
        Gray dashed nodes denote rejected candidate versions.
        Blue arrows denote accepted trunk updates, while orange and green arrows denote accepted side-branch updates using the discovery and refinement selectors, respectively.
        \(\dagger\) marks updates for which AGBU rejects at least one edit.
    }
    \label{fig:spreadsheet_evolution_process}
    \vspace{-1em}
\end{figure}

Figure~\ref{fig:spreadsheet_evolution_process} shows an example of the evolution process. 
The trunk first improves from 16/39 to 21/39, but most of its later
updates are rejected, and its best score is only 22/39. In contrast, the branch path continues to improve and reaches the final best node D:v0030 with a score of 30/39. 
This path uses both discovery and refinement selectors. 
At R:v0007 and D:v0026, AGBU filters out harmful edits while keeping useful ones. This example shows how branch exploration finds a better path when linear evolution stops improving.

\subsection{Validation Results}

\begin{table*}[t]
\centering
\caption{Validation results across benchmarks.
Higher values indicate better performance.}
\label{tab:validation_results}
\footnotesize
\setlength{\tabcolsep}{3pt}
\renewcommand{\arraystretch}{1.08}
\begin{tabularx}{\textwidth}{
c V
c V
*{5}{Y}
}
\toprule
\multirow[c]{2}{*}{\textbf{Model}}
&
\multirow[c]{2}{*}{\textbf{Method}}
&
\multicolumn{5}{c}{\textbf{Validation Score}}
\\
\cmidrule(lr){3-7}
&
&
\textbf{Spreadsheet}
&
\textbf{LiveMath}
&
\textbf{SearchQA}
&
\textbf{OfficeQA}
&
\textbf{TABVERSE}
\\
\midrule
\multirow{7}{*}{
\makecell[l]{DeepSeek-V4\\-flash}
}
& Linear
& 56.41 & 54.29 & 77.00 & 61.22 & 70.00 \\
& Best-of-3
& 71.79 & 57.14 & 77.00 & 65.31 & \textbf{71.43} \\
& Random
& 64.10 & 54.29 & 77.00 & 67.35 & 70.00 \\
& SkillOpt
& 71.79 & \textbf{60.00} & 78.00 & 57.14 & 70.00 \\
& SkillOpt-Lite
& 71.79 & 45.71 & 76.00 & 53.06 & 68.57 \\
& Hermes-SE
& 61.54 & 57.14 & 76.50 & 69.39 & 67.14 \\
\rowcolor{bestrow}
\cellcolor{white}\strut
& \textbf{\workname}
& \textbf{76.92} & \textbf{60.00} & \textbf{82.00} & \textbf{71.43} & \textbf{71.43} \\
\midrule
\multirow{7}{*}{
\makecell[l]{Qwen3.8\\-Flash}
}
& Linear
& 69.23 & 37.14 & 77.50 &  67.35 & 71.43 \\
& Best-of-3
& \textbf{71.79} & 40.00 & 77.50 & 69.39 & 71.43 \\
& Random
& 69.23 & 48.57 & 77.50 & 67.35 & 71.43 \\
& SkillOpt
& 69.23 & \textbf{54.29} & \textbf{78.00} & 59.18 & 67.14 \\
& SkillOpt-Lite
& 66.67 & 34.29 & 77.00 & 69.39 & 68.57 \\
& Hermes-SE
& 61.54 & 51.43 & 77.00 & 69.39 & 67.14 \\
\rowcolor{bestrow}
\cellcolor{white}\strut
& \textbf{\workname}
& \textbf{71.79} & 48.57 & \textbf{78.00} & \textbf{73.47} & \textbf{72.86} \\
\bottomrule
\end{tabularx}
\end{table*}

Table~\ref{tab:validation_results} reports the validation scores of the best nodes used in Table~\ref{tab:main_results}.
\workname achieves the highest validation score, including ties, in nine of the ten model–benchmark combinations.
The only exception is LiveMath with Qwen3.8-Flash.
This is because the validation set of LiveMath is small and agent rollouts are stochastic, making the final score more sensitive to individual samples.
Different benchmark difficulties and data characteristics leave different amounts of room for improvement.
For example, \workname shows larger gains on Spreadsheet and OfficeQA and smaller gains on SearchQA.
Nevertheless, \workname consistently finds nodes with the highest validation scores across these tasks, demonstrating its broad applicability.